\documentclass[sn-mathphys-num]{sn-jnl}

\usepackage{graphicx}%
\usepackage{multirow}%
\usepackage{amsmath,amssymb,amsfonts}%
\usepackage{amsthm}%
\usepackage{mathrsfs}%
\usepackage[title]{appendix}%
\usepackage{xcolor}%
\usepackage{textcomp}%
\usepackage{manyfoot}%
\usepackage{booktabs}%
\usepackage{algorithm}%
\usepackage{algorithmicx}%
\usepackage{algpseudocode}%
\usepackage{listings}%
\usepackage{subcaption}
\usepackage{colortbl}
\usepackage{xcolor}
\usepackage{tabularray}
\usepackage{tabularx}
\usepackage{enumitem}
\newcommand{\OUT}[1]{}

\begin{document}

\title[Multimodal Reinforcement Learning for Robots Collaborating with Humans]{Multimodal Reinforcement Learning for Robots Collaborating with Humans}


\author*[1]{\fnm{Afagh} \sur{Mehri Shervedani}}\email{amehri2@uic.edu}

\author[1]{\fnm{Siyu} \sur{Li}}\email{sli230@uic.edu}

\author[2]{\fnm{Natawut} \sur{Monaikul}}\email{monaiku1@uic.edu}

\author[3]{\fnm{Bahareh} \sur{Abbasi}}\email{bahareh.abbasi@csuci.edu}

\author[2]{\fnm{Barbara} \sur{Di Eugenio}}\email{bdieugen@uic.edu}

\author[1]{\fnm{Miloš} \sur{Žefran}}\email{mzefran@uic.edu}

\affil*[1]{\orgdiv{Department of Electrical and Computer Engineering}, \orgname{University of Illinois Chicago}, \orgaddress{\city{Chicago}, \state{IL}, \country{USA}}}

\affil[2]{\orgdiv{Department of Computer Science}, \orgname{University of Illinois Chicago}, \orgaddress{\city{Chicago}, \state{IL}, \country{USA}}}

\affil[3]{\orgdiv{Computer Science Department}, \orgname{California State University Channel Islands}, \orgaddress{\city{Camarillo}, \state{CA}, \country{USA}}}

\abstract{Robot assistants for older adults and people with disabilities need to interact with the users in collaborative tasks. The core component of these systems is an interaction manager whose job is to observe and assess the task and infer the state of the human and their intent to choose the best course of action for the robot. Due to the sparseness of the data in this domain, the policy for such multimodal systems is often crafted by hand; as the complexity of interactions grows, this process is not scalable. This paper proposes a reinforcement learning (RL) approach to automatically generate the robot's multimodal policy. In contrast to the traditional dialog systems, our agent is trained with a simulator that uses human data and can deal with multiple modalities, such as language and physical actions. We use a simple high-level reward function that needs no fine-tuning and enforce some preconditions to speed up the training process. We conducted a human study to evaluate the performance of the system in the interaction with a user. Our designed system shows promising preliminary results when it is used by a real user.}


\maketitle
\section{Introduction}
\label{sec: Introduction}
\begin{figure}[t]
\centering
\includegraphics[width=\columnwidth]{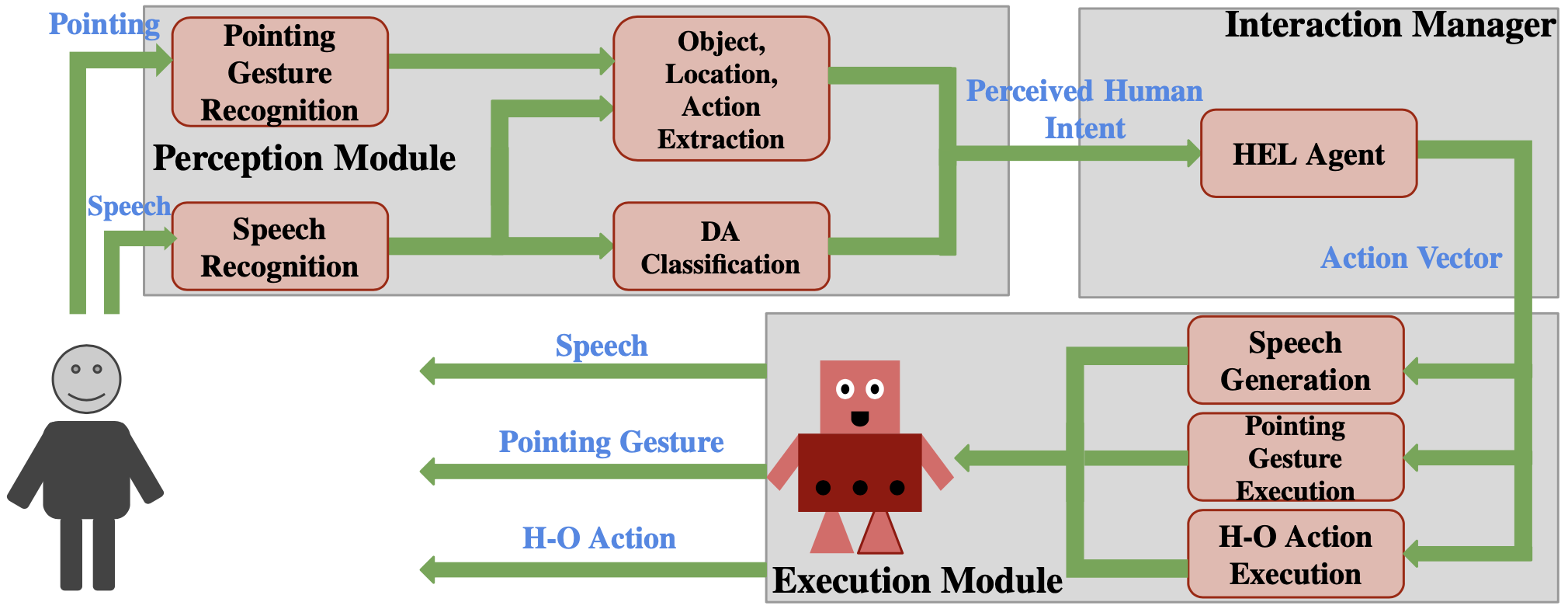}
\caption{The \emph{Sense-Plan-Act} cycle in an assistive robot}
\label{fig: sense-plan-act}
\end{figure}

Assistive robots built to help older people and people with disabilities with activities of daily living (ADLs) such as cooking or cleaning must be able to work in tandem with their users to complete their tasks. These interactions are typically multimodal (involving speech, gestures, and physical actions) and collaborative (requiring some level of engagement from both participants). As with other autonomous robots, these assistive robots can be decomposed into three main components that perform the typical \textit{sense-plan-act} cycle throughout the task: a perception module that collects and processes sensory data (potentially from multiple modalities) to understand the environment and the user's actions; an interaction manager that then determines an appropriate response for the robot; and an execution module that enables the robot to perform this action (Fig.~\ref{fig: sense-plan-act}).

One way to design an effective interaction manager for such a robot is to study how two humans interact in collaborative tasks. We previously proposed a novel architecture called a Hierarchical Bipartite Action-Transition Networks (HBATNs) for multimodal human-robot interaction management~\cite{8968505, monaikul2020role}, which was developed largely based on our corpus of interactions between elderly individuals (labeled ELD) and nursing students (labeled HEL) assisting them in completing ADLs~\cite{chen2015roles}. In particular, we targeted what we call the \textit{Find} task, in which two participants collaborate to locate an object that is not visible in the environment. We observed participants in this task interacting via speech, pointing gestures, and \textit{haptic-ostensive} (H-O) actions~\cite{chen2015roles} that bring physical objects into conversational focus through touch.

We showed that HBATNs could effectively provide a policy for the robot. But while the construction of the HBATNs for the \emph{Find} task was completely grounded in our human-human interaction data, it was ultimately crafted by hand, and it would need to be redesigned for new tasks. On the other hand, recently, researchers have been utilizing Large Language Models (LLMs) to integrate common sense knowledge into robotics. These developments enable robots to learn complex policies that require extensive background knowledge and semantic understanding~\cite{ouyang2022training, brown2020language, achiam2023gpt}. However, this approach demands a large amount of data and presents significant challenges, particularly in grounding LLMs for multimodal tasks~\cite{brohan2023can, driess2023palm}.

This paper proposes a more scalable interaction manager compared to HBATNs, employing reinforcement learning (RL) to automatically
generate a policy for a robot engaged in a collaborative task. The approach addresses the issues of limited data and grounding encountered in LLM-based methods. With RL, the robot learns to differentiate desirable actions from undesirable actions through trial and error with feedback from its interactive environment in the form of rewards and penalties~\cite{sutton2018reinforcement,sallans2004reinforcement}. We demonstrate the effectiveness of this RL-based interaction manager by training the robot's agent to assume the HEL role, collaborating seamlessly with a human who takes on the ELD role during the \textit{Find} task. 

A major challenge in training this agent is building an interactive environment that can provide the agent with rewards. Here, the environment includes not only the physical space in which the agent acts but also the human user acting as ELD. To be able to perform RL at scale, we developed a simulated environment with a neural network-based user simulator~\cite{10309444} that was inspired by Behavioral Cloning~\cite{bratko1995behavioural, torabi2018behavioral}. This user simulator is trained on \emph{Find} task data from our ELD-HEL interaction corpus. Additionally, owing to the relatively limited size of our corpus for RL-based training and the inherent challenges in interactions involving a robot, which are more prone to error than those between two humans due to imperfect sensors and algorithms, we augmented our data with synthetic, data-driven misunderstandings and equipped our environment with an error injection module. 
 
Our main contribution is an interpretable RL-based interaction manager for multimodal collaborative robots. The reward function is simple, and some preconditions are enforced to prevent unwanted policy searches and speed up the training process. The end-to-end user simulator provides an RL training environment. 
While training the agent through RL, it is expected that mistakes will be made by the agent, such as misinterpreting some states. The user simulator can provide proper responses in those cases~\cite{10309444}. 

Furthermore, we designed and conducted a
user study involving human subjects. We developed and implemented the necessary components of the perception and execution modules and evaluated the performance of both the RL-based interaction manager and each individual component of the perception and execution modules. The user study showed that the robot agent achieved high accuracy and user satisfaction. The novel approaches we employed to design and integrate the components, along with the remarkable performance of the RL-based interaction manager, contributed to these positive results.

The remainder of the paper is organized as follows. Section~\ref{sec: Related Work} reviews related literature, Section~\ref{sec: Simulator} presents our user simulator, Section~\ref{sec: Framework} outlines our RL framework, Section~\ref{sec: Evaluations} describes the user study we carried out to evaluate the performance of our learned interaction manager, and Section~\ref{sec: Discussion} discusses the advantages of our system compared to~\cite{8968505}.

\section{Related Work}
\label{sec: Related Work}

Machine learning has become an indispensable tool in robotics. The critical aspect of interaction with the environment makes RL a good candidate for learning a policy from sequential human-robot interaction data. After theoretical guarantees for Q-learning were published in the '80s~\cite{watkins1992q}, RL algorithms have developed greatly, including Deep-Q-Learning, Actor-Critic, and Trust Region Policy Optimization~\cite{hester2018deep, mnih2016asynchronous, schulman2017proximal}. For example, in \cite{mnih2016asynchronous}, the authors demonstrate how RL agents can outperform humans in the Atari game environment. Inverse RL is also popular because when data is limited, human guidance can significantly reduce the set of states that need to be explored. As an example, in~\cite{woodworth2018preference}, the authors introduce a preference-inference Inverse-RL model for assistive robots that learn different preferences by observing users performing tasks. In this work, we are interested in learning an agent for an intelligent assistive robot, and to that end, we propose a Deep-Q-Learning (DQL) framework with a DA{\footnotesize GGER} warm-up~\cite{ross2011reduction} which helps us tune the RL reward function more efficiently.

In RL, a mandatory component to complete the feedback loop for training is a simulated user or an interactive environment. In human-robot interaction, the simulated user (which mimics humans) should be reliable but should also generate a variety of actions to ensure RL exploration. The authors of~\cite{thomaz2006reinforcement} examine whether the human-given reward is compatible with the traditional RL reward signal.  
In~\cite{sharkawy2021human}, the authors assume that the human is an expert in interacting with the robot; that is, humans will always give actions designed to be easily understood by the robot. In this case, the human policy is a fixed and deterministic function of the robot's movements. In our case, this expert assumption for the user simulator is inaccurate since the user, who might have a disability and need help, cannot be seen as an expert. Instead of giving the best decision, our proposed user simulator can capture the user's decision preference.
One can personalize the user simulator by training it based on their own collected data. Our RL agent can always successfully assist these personalized users by acquiring information or taking the initiative. 

Several works exist that propose RL frameworks for training interactive dialogue systems to assist users~\cite{liu2018dialogue, chen2017survey, zhao2016towards, li2017end}. For example, in~\cite{li2017end}, a Deep Q-learning network architecture is proposed for a dialogue system to help humans access information or complete tasks. The agent is trained with a language generation machine as the user simulator. One of the main differences between this work and ours is that we work with actual human data to develop our user simulator. In addition, in dialogue systems, the agent is limited to a single modality (language), while we deal with a multimodal system, which makes it more complex. Another related work is~\cite{park2019model} in which Q-learning is employed to learn a personalized policy based on children's facial reactions and pose data in the class. Although they work with actual human data, they are only concerned with visual data from poses and faces captured by a camera. Throughout the recent literature, not much work considers a multimodal setting with gestures, physical actions, and dialogue while training the intelligent agent based on actual human data. 

In this paper, we build on the HBATN framework proposed in~\cite{8968505}, which allowed the agent to manipulate the objects in the environment and interact with users through pointing gestures, haptic-ostensive (H-O) actions~\cite{chen2015roles}, and speech. This data-driven model was developed based on our previously collected corpus of human-human interactions~\cite{chen2015roles}. The task studied in detail was the \textit{Find} task, an interaction scenario in which a human and a robot work together to find an object of interest in the environment. HBATNs model both agents simultaneously to maintain the state of a task-driven multimodal interaction and plan the subsequent robot moves. Although this model provides an interpretable decision-making process (rule-based and constructed by hand), it lacks scalability. Also, extracting the policy manually is a daunting task. One of our main contributions in this paper is to automate this policy extraction process using RL.  

\section{User Simulator}
\label{sec: Simulator}
A critical component of the RL training cycle is the environment in which the agent to be trained can perform actions and receive feedback. In a collaborative setting, the environment also includes the partner with whom the agent interacts. As illustrated in Fig.~\ref{fig: sense-plan-act}, our HEL agent decides on its next action based on the perceived human intent. As a result, we need to provide the HEL agent with this input in the RL training cycle as well.

We focus on the \textit{Find} task, for which we have annotated human-human interaction data. To train an agent to perform the role of HEL in the \textit{Find} task using RL, we thus require a user simulator that can act as an ELD and provide the same kind of response we observed in our data. 
Previously, an HBATN model was developed to represent the states and actions of both participants in the \textit{Find} (and more generally, a collaborative) task~\cite{8968505}. The task was decomposed into subtasks with the goal of identifying a type of target object $O_T$, identifying locations $L$ potentially containing the object, and, ultimately, locating the target object $O$. In~\cite{monaikul2020role}, we showed that our HBATN model, equipped with a trained classifier that determines which subtask the participants are currently in, can model and perform both HEL and ELD behavior; however, training an agent using RL requires the agent to be able to explore its space of states and actions, some of which are never seen in our data, which can lead to erratic behavior that the subtask classifier was not trained to handle.

Moreover, the number of trials required to perform RL grows significantly with the size of the agent's search space, and putting a human user in the training cycle to provide the perceived human intent input to the RL agent is not realistic. Thus, we developed our user simulator. We initially built the skeleton of our RL framework and demonstrated its robustness by conducting a preliminary user study. However, the preliminary results showed that there was still room to improve the user simulator and, as a result, improve the RL policy. In~\cite{10309444}, we describe the details of the user simulator and investigate the factors leading to better performance. A summary of the improved user simulator that is used for training the RL agent in this work is presented here. 






\subsection{Feature Extraction}

The initial step in designing the user simulator (ELD) is to determine what input is needed for the simulator so it can output an appropriate response to the agent (HEL) action. We consider three main parameters: (1) ELD's belief of HEL's knowledge of $O_T$, which can be one of three values (ELD believes HEL does not know the target $O_T$, ELD believes HEL does know the target $O_T$, or ELD believes HEL is thinking of a different $O_T$), (2) ELD's belief of HEL's knowledge of $L$, and (3) ELD's belief of HEL's knowledge of $O$. We supplement these with additional features representing HEL's action that ELD responds to: (4) what or where HEL pointed to, if a pointing gesture was performed, (5) what H-O action HEL performed and on what, if an H-O action was performed, and (6) the HEL's action and the \emph{Dialogue Act} (DA) -- roughly, intent indicated through speech -- of HEL's utterance, which we have extensively studied and built classifiers for~\cite{chen2015roles, 8968505, monaikul2020role}, (7) previous ELD's action and DA.

We assume that both ELD and HEL have a finite set of actions. We developed a list of actions for each role based on the \textit{Find} task corpus so that HEL's action label could be input to the user simulator and so that the user simulator can generate its action labels as output. ELD actions include providing the $O_T$ or the $L$ and giving an affirmative or negative response, while HEL actions include requesting the $O_T$ or the $L$ and verifying a potential $O_T$, $L$, or $O$. 

We designed our user simulator to output ELD's next action label, the DA tag associated with ELD's action (to simulate speech recognition and classification when the agent interacts with a human user), and the updated state representation of ELD's beliefs of HEL's current knowledge of the task. The action label and DA tag can then be passed to the HEL agent during RL training, while the updated state representation of ELD's beliefs gets passed to the user simulator at the subsequent step. When interacting with a human user, the HEL agent will need to be equipped with classifiers to determine the user's action label and DA-given sensory input. We elaborate on these components in Section~\ref{subsection: Implementation}.

\subsection{Data Annotation}
We turn to the \textit{Find} task data available in the ELDERLY-AT-HOME corpus~\cite{chen2015roles}. This data was previously transcribed and annotated for DAs, pointing gestures, and H-O actions. To train our user simulator with this data, we supplied additional annotations around ELD beliefs of HEL's knowledge of $O_T$, $L$, and $O$, and ELD and HEL actions using our list of actions. Details on the data annotation approach can be found in~\cite{10309444}.

\subsection{Data Augmentation}
The \textit{Find} task data provided a strong foundation for training our user simulator, but there were no instances in which ELD believed HEL had the wrong $O_T$ or $L$ in mind. The data also contained very few instances in which ELD believed HEL did not know $O_T$. Only in the beginning of each trial did this occur; after ELD provided the $O_T$, HEL never needed to ask for $O_T$ again. 

The lack of these states is unsurprising given that the interactions were between humans who could hear and see each other well. However, when we train our HEL agent with RL, we expect that the agent will make mistakes, misinterpreting or not even interpreting $O_T$ or $L$, so the user simulator needs to be able to respond appropriately. We, therefore, augment our data with synthetic but grounded examples that cover these missing or infrequent states.



Details of the data augmentation approach with corresponding examples can be found in~\cite{10309444}.

\subsection{Model Architecture and Training}

Our model is a neural network consisting of three fully connected (dense) layers, and a dropout layer (ratio=0.2) to prevent overfitting and improve the ability of the model to generalize better. We utilized the Cross-Entropy loss function and Adam optimizer during training. The training process lasted for a maximum of 100 epochs, but we also evaluated the model's performance on the validation set while training to allow for early stopping. The inputs to the neural network were the features described earlier, while the outputs were the ELD's next state, dialogue act, and action, which were manually annotated in the data. We implemented the model using the PyTorch library~\cite{NEURIPS2019_9015}.

In addition to overall accuracy, we evaluated the model on the classification accuracy of each individual output of the model, i.e., the classification accuracy for (1) the predicted belief ELD holds of HEL's knowledge of $O_T$, $L$, and $O$; (2) the predicted DA; (3) the predicted action. The accuracy results are provided in Table~\ref{table: user-sim}.

\begin{table}[h]
\caption{Classification Accuracy of User Simulator~\cite{10309444}}
\label{table: user-sim}
\begin{tabular}{@{}cccc@{}}
\toprule
Overall Accuracy & Action Accuracy & DA Accuracy  & State Accuracy  \\
\midrule
0.85\% & 77.89\% & 75.38\% & 89.44\% \\
\bottomrule
\end{tabular}
\end{table}

\section{Reinforcement Learning Framework}
\label{sec: Framework}

\begin{figure}[t]
\centering
\includegraphics[width=\columnwidth]{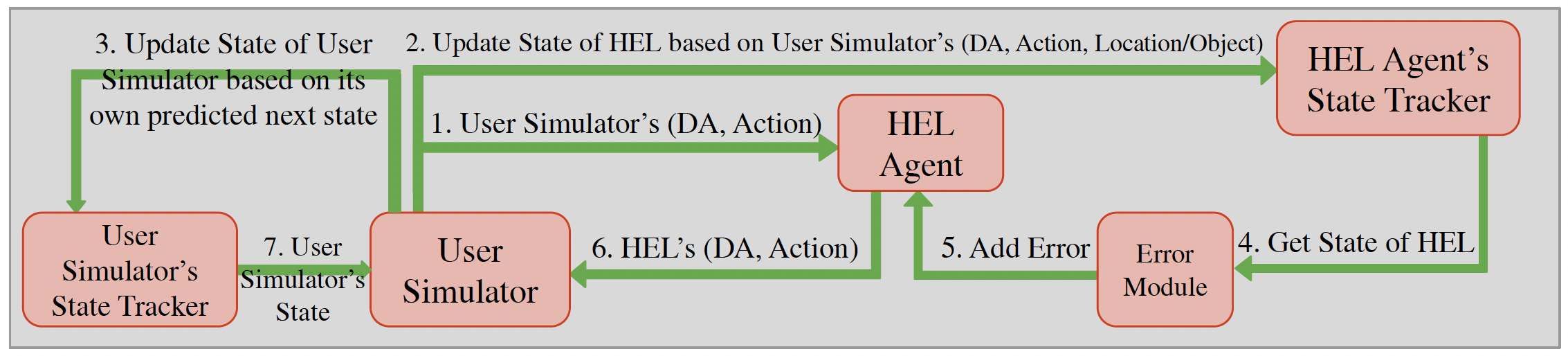}
\caption{The interaction between the User Simulator and the HEL agent during RL}
\label{fig: arch_IL_RL}
\end{figure}

In this RL problem, we employ a Deep-Q-Network (DQN)~\cite{mnih2013playing} model to act as our assistive robot (HEL) agent by having it interact with the interaction environment containing the user simulator proposed in the previous section. To enhance the training, we warm up the agent before starting the training loop using DA{\footnotesize GGER}, an Imitation Learning (IL) algorithm~\cite{ross2011reduction}. The warm-up phase happens before initiating the RL training cycles. This step is crucial due to the fact that learning the appropriate agent behaviors with a simple reward function is extremely difficult and training such an agent from scratch is tedious.

The flow of the interactions between the HEL agent and the user simulator while training the HEL agent is illustrated in Fig.~\ref{fig: arch_IL_RL}. The state of the HEL agent consists of three variables: the state of HEL's \emph{$O_T$}, HEL's \emph{$L$}, and HEL's \emph{$O$}. The possible values for each variable are 0, 1, or 2. The variable is encoded with 0 when it has not been determined yet, with 1 when it matches the user simulator’s belief about it, and with 2 when there is a mismatch between the HEL value and the user simulator’s belief.

As depicted in Fig.~\ref{fig: arch_IL_RL}, the HEL agent takes the user simulator's action and the previous HEL's state as input and outputs a \emph{Dialogue Act} (DA) tag and an action vector encoding the HEL's utterance and physical action, respectively.

\subsection{Model Architecture}
The HEL agent network includes two fully connected layers followed by a dropout layer (ratio=0.1). Then, the output of the dropout layer is fed to the output layer, followed by ReLU activation, and a vector encoding HEL's (DA, action) output pair is given. For implementation, we used the PyTorch library~\cite{NEURIPS2019_9015}.

\subsection{DA{\footnotesize GGER} Warm-up}
\label{subsec: Framework-Dagg}
First, we have our HEL agent interact with the user simulator and run the DA{\footnotesize GGER} algorithm as a warm-up stage. We don't let the agent get fully trained; we only use it to obtain a good initial guess for the subsequent RL training. Using DA{\footnotesize GGER} only, without subsequent RL training, results in poorer performance of the HEL agent.

The user simulator initiates the interaction, the state of the HEL agent is updated according to the received input, and an action is picked according to its current state and the user simulator's action. An error module is deployed before passing the HEL's state to the HEL agent.

The error module is necessary due to the fact that in the data when $O_T$/$L$/$O$ information is given to the HEL by the ELD, most of the time, the state corresponding to that changes to 1, i.e., the HEL's understanding of $O_T$/$L$/$O$ is the same as what human has uttered. However, we would like to generalize the framework better so that it also covers the states when the HEL's understanding of $O_T$/$L$/$O$ does not match the human's utterances. Thus, the error module in 25\% of the cases where HEL's understanding of $O_T$/$L$/$O$ is 1 changes that to 2.

For training the HEL agent using the DA{\footnotesize GGER} algorithm, we need to run the algorithm for $N$ (here, $N=25$) iterations. Here, we call each iteration one \emph{Episode} since the interactions between the ELD and the HEL are defined to be episodic. During each episode, one entire interaction consisting of at most $M$ (here, $M=25$) turns between the HEL agent and the user simulator takes place. That interaction is successful if the HEL agent finds the object the user simulator requested before reaching turn $M$. Otherwise, the interaction is unsuccessful.

During each episode, the HEL agent executes the current learned policy. Throughout execution, at each turn, the expert's action, which we get from the roll-outs extracted from our ELDERLY-AT-HOME data, is also recorded but not executed. After sufficient data is collected, it is aggregated together with all previously collected data. Eventually, the cross-entropy algorithm generates a new policy by attempting to optimize performance on the aggregated data. This process of executing the current policy, correction by the expert, and data aggregation and training is repeated.

\begin{figure}[h]
\centering
\begin{subfigure}[b]{0.48\columnwidth} 
\centering
\includegraphics[width=\columnwidth]{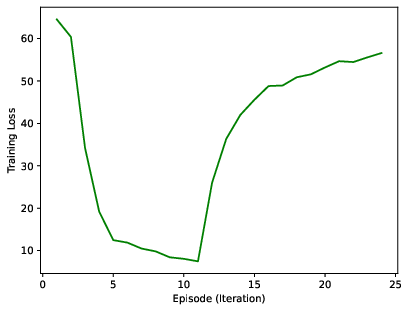}
\caption{{\scriptsize Training Loss}}
\label{fig: dagger_loss}
\end{subfigure}
\hfill
\begin{subfigure}[b]{0.51\columnwidth} 
\centering
\includegraphics[width=\columnwidth]{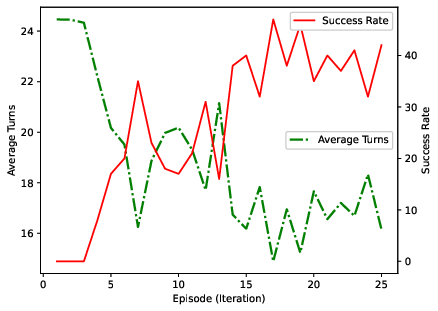}
\caption{{\scriptsize Success Rate and Average Turns}}
\label{fig: dagger_s_rate}
\end{subfigure}
\caption{DA{\footnotesize GGER} Algorithm Training Evaluations}
\label{fig: dagger_evaluations}
\end{figure}

Fig.~\ref{fig: dagger_evaluations} shows training loss, success rate, and average turns during each episode for DA{\footnotesize GGER} training on the agent for 25 episodes. However, as mentioned before, we do not want a fully DA{\footnotesize GGER} trained HEL agent. Therefore, the HEL agent's network weights are saved at episode 10 (a good mid-point where the loss is not minimized and the success rate is not maximized yet) to be used later on for training the HEL agent using the DQL algorithm. As seen in Fig.~\ref{fig: dagger_evaluations}, overfitting occurs after a few episodes due to a very limited amount of data. This highlights the significant role of RL in training our interaction manager.

\subsection{Deep-Q-Learning}
For training our HEL agent in the RL cycle, we need to initialize a \emph{Policy Network} and a \emph{Target Network}. The former network's weights will be optimized for obtaining the optimal policy, and the latter network will be used to track the target Q-values associated with each individual action for input states~\cite{mnih2013playing}. 
The policy we extracted at episode 10 of DA{\footnotesize GGER} training is used to initialize the weights of the \emph{Policy Network} and the \emph{Target Network} for DQL. The model architectures for the \emph{Policy Network} and the \emph{Target Network} are exactly the same as the model used previously for DA{\footnotesize GGER} training due to the fact that we simply copy the weights extracted from DA{\footnotesize GGER} network into these two networks. This makes running the DQL algorithm on our RL framework much more efficient in terms of time and space. 

Another advantage of warming up the HEL agent by DA{\footnotesize GGER} training is that by pushing the weights of the policy toward the expert's policy, we don't need to hard code complex human behavior in the \emph{Reward Function}. A simple \emph{Reward Function} combined with DA{\footnotesize GGER}-half-trained HEL agent makes the DQL algorithm on our HEL agent run much faster. 

For our reward function, we considered a small negative reward, $-r$ (here, $r=1$), for each HEL agent's move. This motivates the agent to finish the task sooner rather than later. If the interaction is unsuccessful and arrives at some latest allowed turn $M$ (here, $M=20$), the transition is penalized by $-2 \, r$. If the interaction successfully ends before reaching turn $M$, that transition is rewarded as $2\, r$. We also set some ground rules as \emph{Preconditions}. The \emph{Preconditions} are as follows: (1) the HEL agent must not take the action of verifying $O_T$ before $O_T$ is uttered by the user simulator; (2) the HEL agent must not take the action of verifying $L$ before $L$ is uttered by the user simulator; (3) the HEL agent must not take the action of verifying $O$ before both $O_T$ and $L$ are uttered by the user simulator. If any of the \emph{Preconditions} are violated, that action is penalized with a large negative number $-Z$ (here, $Z=50$).

The interactions between the HEL agent and the user simulator while running the DQL algorithm also follow the flow in Fig.~\ref{fig: arch_IL_RL}. At each turn, a tuple of \emph{(state, reward, action, next state)} is stored in the memory. Every $C$ episodes (here, $C=500$), the weights of the \emph{Policy Network} are optimized using the \emph{Mean-Squared-Error} cost, and for every $m \, C$ episodes (here, $m=4$), the weights of the \emph{Policy Network} are copied into the \emph{Target Network}. 

\begin{figure}[h]
\centering
\begin{subfigure}[b]{0.503\columnwidth} 
\centering
\includegraphics[width=\columnwidth]{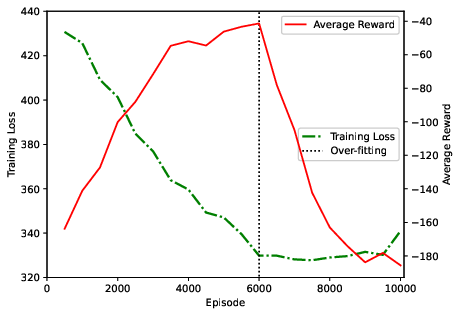}
\caption{{\scriptsize Training Loss and Average Reward}}
\label{fig: dql_loss}
\end{subfigure}
\hfill
\begin{subfigure}[b]{0.49\columnwidth} 
\centering
\includegraphics[width=\columnwidth]{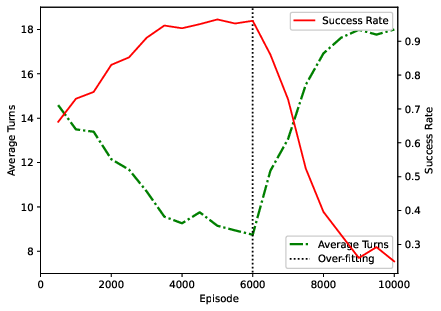}
\caption{{\scriptsize Success Rate and Average Turns}}
\label{fig: dql_s_rate}
\end{subfigure}
\caption{DQL Algorithm Training Evaluations}
\label{fig: dql_evaluations}
\end{figure}

Fig.~\ref{fig: dql_evaluations} shows training loss, average reward, success rate, and average turns at each episode for the full DQL algorithm training of the HEL agent. The reason that average turns start to increase and the success rate gradually decreases after about episode 6000 is that the memory of the agent has been filled up, so part of the memory is emptied and starts to get filled with new samples. Catastrophic forgetting has happened at this point (the agent has been over-fitted). We thus use the policy obtained just before this phenomenon occurs.

\section{Experimental Evaluations}
\label{sec: Evaluations}
To evaluate our framework in the real world, we designed a robot implementation that allows humans to interact with the HEL agent obtained through RL. In the following section, we explain the details of our interface and the evaluation procedure.

\subsection{Robot Implementation}
\label{subsection: Implementation}
Since our framework is trained on multimodal human data, it is capable of interpreting and executing multimodal actions, particularly language and physical actions. We implemented our framework on the Baxter robot from Rethink Robotics since it can perceive and perform multimodal actions.

We developed and implemented a robot implementation following the architecture of the Perception Module and the Execution Module shown in Fig.~\ref{fig: sense-plan-act}. Since the HEL agent (RL policy) is trained based on DA tags and the action vectors generated by the user simulator, it expects to receive the human subject's DA tags and action vectors during the test phase as well. Consequently, we need to extract the DA tags and action vectors from human utterances and physical gestures. Therefore, we have included speech-to-text, action extractor, and DA classifier components, which will be explained in detail later on. Additionally, we need pointing gesture, H-O action, and speech generation components for the agent to respond to the human. We will also elaborate on the implementations of these components.

\subsubsection{Perception Module}
Human speech is passed to the Google Cloud Speech-to-Text API~\cite{s2t_api} for speech recognition. The transcribed text is then passed to a DA classifier and an action extractor. We developed our DA classifier by deploying ALBERT~\cite{lan2019albert} and fine-tuning it using our ELDERLY-AT-HOME data. To enhance the generality and accuracy of the classifier, we augment the data with synonym replacement, random insertion, random deletion, and random swap techniques. Since ALBERT includes its own \textit{AlbertForSequenceClassification} model, we used this model without any modifications. We set the number of classes to correspond with the number of DA tags in our classification task. We used the gold-standard DA tags as ground-truth labels and optimized the weights using the cross-entropy loss function and the Adam optimizer. Our ALBERT-based DA classifier achieved an accuracy of 88.41\% when tested on 10\% of the data. 

Our action extractor component works based on the objects and locations it extracts from human utterances and pointing gestures. For example, if it detects one of the objects in the utterance, that's labeled as \emph{``Give $O_T$"} action; if it detects one of the locations in the utterance and/or in the pointing gesture, that's labeled as \emph{``Give $L$"} action; if it detects one of the objects in the utterance as well as one of the locations in the utterance and/or in the pointing gesture, that's labeled as \emph{``Give $O_T, L$"} action. To extract the words of interest (objects and locations present in the experiment), we built a dictionary based on the NLTK dictionary~\cite{bird2009natural}, including all the words that could be pronounced similarly or close to our target words. This helps compensate for speech recognition errors where similarly sounding words that do not make sense in context might be returned.

As mentioned above, the human could use pointing gestures to perform \emph{``Give $L$"} and \emph{``Give $O_T, L$"} actions. To detect those pointing gestures, we deployed a pre-developed model from~\cite{8968505} and calibrated it for our experimental environment. 

Finally, the Perceived Human Intent vector made of the $(O_T, L, O)$ state, the DA tag, and the action tag is passed to the Interaction Manager, the RL policy.

\subsubsection{Execution Module}

The Interaction Manager produces an action vector based on which the robot performs its next move. As depicted in Fig.~\ref{fig: sense-plan-act}, this action vector is fed into each component of the execution module to execute its specific task. For instance, the robot can point by adjusting its arm to predetermined positions near the target area. The robot can open a drawer based on a pre-programmed trajectory for haptic interactions. Additionally, the robot's display screen can show an image captured by its built-in hand camera, mimicking the action of visibly presenting an object to the user.

We first need to transform the DA tags encoded in the produced action vector into sentences for speech generation. To achieve this, we developed a rule-based text generator that is based on the ELDERLY-AT-HOME corpus. Then, we used the Pyttsx3 Python library~\cite{pyttsx3} to convert the agent's sentences into speech. Finally, the Baxter robot performs physical actions while speaking the generated speech.

\subsection{User Study}

We conducted a user study with 12 healthy adults (6 women and 6 men) who were recruited to interact with the robot. Each subject performed 5 to 7 trials, resulting in a total of 75 trials. 

The experiment environment was a room equipped with a drawer, a shelf, and a cabinet. Users could choose between red, green, and yellow cups, as well as red, green, yellow, and white balls. At the beginning of each trial, objects were randomly placed to different locations. Although users were aware of the available locations and objects in the room, they did not know the specific location of each item. Subjects were instructed to choose the object of interest at the beginning of the trial and guide the agent through different locations to find that object.

The data collected in this user study will be added to our publicly available ELDERLY-AT-HOME corpus. 

\bigskip
\textbf{Ethical approval declarations}

\begin{enumerate}
\item Approval Statement: All experimental protocols were approved by the University of Illinois Chicago Institutional Review Board (IRB).

\item Accordance Statement: The methods described in this study were carried out in accordance with the guidelines and regulations provided by the University of Illinois Chicago Institutional Review Board (IRB).

\item Informed Consent Statement: Informed consent was obtained from all participants involved in experiments using the University of Illinois Chicago Institutional Review Board (IRB) Social, Behavioral, and Educational Research Informed Consent Template.
\end{enumerate}

\subsection{Quantitative Results}
\label{subsection: quant-results}

We evaluated the performance of the system by measuring the accuracy of each component individually, along with assessing the overall system quality. In addition, we compared the performance of the system with the performance of the system presented in~\cite{8968505} as a baseline.

To evaluate the overall performance of the entire system, we calculate and report the percentage of the HEL's non-eligible moves in response to the human's move (speech and action). A non-eligible move is one where the agent's response is not consistent with the human's previous moves. For example, if the human says: ``Please get a cup." and the agent responds with: ``Did you say inside the cabinet?" while the human has not given any information about the location yet, this is a non-eligible move. In other words, non-eligible moves are those actions that do not make sense to the human participants in the study based on the speech they have uttered and the physical actions they have taken so far. Non-eligible moves could result from speech recognition errors, DA classification errors, pointing gesture recognition errors, and/or wrong actions by the HEL agent (RL policy) itself. There were 469 total inputs (human moves) to the system, and 46 non-eligible responses occurred. This means that 9.8\% of the total actions taken by the HEL agent were non-eligible. Non-eligible moves could be equivalent to non-understandings in~\cite{8968505}.

To evaluate speech recognition, we calculate the percentage of human moves in which Serious Speech Recognition Errors (SSREs) occurred. The definition of SSRE is consistent with~\cite{8968505}: ``An SSRE is a speech recognition error that alters the utterance in such a way that a subtask could not be successfully completed." We remind the reader that subtasks (we talked about subtasks at the beginning of Section~\ref{sec: Simulator}) are still present in our case; they are implicitly represented by the RL policy.
We found that SSREs occurred in 6.4\% of human moves.

To evaluate DA classification, the percentage of human moves in which wrong DAs occurred is calculated. The wrong DAs were present in 15.14\% of human moves. For the pointing gesture recognition component, we computed the percentage of pointing gestures performed that were either not recognized or not tagged with the correct intended location. We determined that 13.64\% of the pointing gestures were detected incorrectly. 

To evaluate the Interaction Manager itself (HEL agent/RL policy), we took three approaches. First, we investigated the contribution of the Interaction Manager to non-eligible moves. Thus, we calculated the percentage of the robot's moves without SSREs, DA errors, pointing gesture recognition errors, and instances where RL policy produced non-eligible moves. Out of 469 human inputs to the system, 370 were detected with no errors. Among these, the response to 7 inputs resulted in non-eligible moves. This corresponds to a real-time accuracy of 98.11\% for the RL policy. As mentioned before, there were 46 non-eligible robot moves in total, meaning that only 7 out of 46 (15.22\%) of non-eligible moves were caused by the RL policy. Thus, 36 non-eligible moves (84.78\%) are due to SSREs, wrong DAs, wrong pointing gestures, or simply non-eligible moves taken by the human subject (humans could make mistakes in a natural interaction, too). The definition of a non-eligible move by the human subject is consistent with the definition of a non-eligible move by the HEL agent.

In the second approach, we calculated the percentage of the HEL's (RL policy's) wrong moves with respect to the human's detected move (perceived human intent vector). A wrong move is one where the agent's response does not correspond to the previous perceived human intent vector. To check whether a move is wrong, we refer to the ELDERY-AT-Home data. If the perceived human intent vector could be decoded and extracted from the corpus, that's a valid vector. An invalid vector could occur due to SSREs, wrong DAs, or wrong pointing gestures. If the HEL's response to that valid vector matches the one from the corpus, that's a correct move; otherwise, it's a wrong move. There was a total of 436 valid perceived human intent vectors, and the HEL agent responded to 425 of them with correct actions (11 wrong actions). This leads to a $97.47\%$ real-time accuracy of the RL policy, which is consistent with the one reported in the first approach.

In the third approach, we compared the performance of the two Interaction Managers from this study (RL policy) and the one from~\cite{8968505} (HBATNs model). Out of 436 valid perceived human intent vectors, 404 were found in HBATNs. The HEL agent's responses to 395  ($97.77\%$) of those input vectors matched those of HBATNs (only 9 wrong actions). This is still consistent with the real-time accuracy of RL policy obtained in the first and second approaches. Putting together the results from the last two approaches, we see that the RL policy gives correct responses to 30 out of 32 valid perceived human intent vectors that were not found in HBATNs.

We also report the average length of interactions as well as the success rate across all trials. A trial is considered successful if the agent finds the object that the human subject is looking for in fewer than 15 turns. If the agent reaches the 15th turn without locating the object, the trial is deemed unsuccessful. The low rate of non-eligible/non-understanding system responses (9.8\%), coupled with the low number of average turns (12.6 average moves, equivalent to 6.3 average turns, where each turn consists of one move taken by the human and one move taken by the agent), and the high success rate (96\%), underscores the significant advantage of our multimodal RL framework. Each component complements the others, compensating for any flaws and enhancing overall performance.

The quantitative results of this human study are summarized in Table~\ref{table: quant_results}. Comparing the results of our human study with the work in~\cite{8968505}, we see a great improvement in the overall performance of the system. The percentage of non-understandings decreased from 11.7\% to 9.8\%. The average number of moves decreased from 15.6 to 12.6, and the success rate increased from 85.7\% to 96\%. However, since two different groups of subjects were recruited for the two different studies, we investigate this improvement of our system in more depth in Section~\ref{sec: Discussion}.

\begin{table}[t]
\caption{Results of the preliminary user study}\label{table: quant_results}
\begin{tabular*}{\textwidth}{@{\extracolsep\fill}lcccccc@{}}
\toprule
Avg. & Success& Non-eligible & SSREs & Wrong & Wrong & RL Policy \\
\#Turns & Rate & Actions &  & DAs &  Pointing &Acc. \\
\midrule
6.3 & 96\% & 9.8\% & 6.4\% & 15.14\% & 13.64\% & $\approx$97.5\% \\
\bottomrule
\end{tabular*}
\end{table}

\subsection{Survey Results}
\label{subsection: survey-results}
We also evaluated the performance of the framework by asking the participants to fill out a survey upon finishing their experiment. This survey was designed following the guidelines provided in~\cite{hoffman2020primer}. The Likert scales were chosen for the responses inspired from~\cite{schrumConcerningTrendsLikert2023}. The survey consisted of 15 questions listed in Appendix~\ref{sec: Appendix}. We categorize those questions into the following groups: (1) Comfort (Likert 1 to 7); (2) Responsiveness (Likert 1 to 7); (3) Predictability (Likert 1 to 7); (4) Familiarity and Learning (Likert 1 to 5); (5) Fatigue and Habituation (Likert 1 to 5); and (6) Overall Quality (Likert 1 to 5). The average values of the responses for the 12 participants are summarized in Table~\ref{table: survey_results}. 

\begin{table}[h]
\caption{Average responses to the survey}\label{table: survey_results}
\begin{tabular}{@{}lllllll@{}}
\toprule
\scriptsize{Question \#} & G1 & G2 & G3 & G4 & G5 & G6\\
\midrule
\scriptsize{Average Likert} & 5.89 & 5.69 & 6.08 & 4.33 & 1.69 & 4.25\\
\bottomrule
\end{tabular}
\end{table}

\textbf{Comfort} The average Likert of 5.89 for the first group of questions indicates that: 
\begin{enumerate}
    \item Users felt physically and emotionally relaxed while interacting with the system.
    \item Users perceived the robot as a suitable and acceptable assistant for the task.
    \item Users found the interaction with the robot to be natural and smooth.
\end{enumerate}

\textbf{Responsiveness} The average Likert of 5.69 for the second group of questions indicates that:
\begin{enumerate}
    \item Users perceived the robot as attentive and responsive to their needs and instructions.
    \item Users perceived the robot as proactive and capable of autonomously driving the interaction when necessary.
    \item Users perceived the robot's behavior as human-like, making the interaction more intuitive.
\end{enumerate}

\textbf{Predictability} The average Likert of 6.08 for the third group of questions suggests that: 
\begin{enumerate}
    \item Users had high levels of trust in the robot's capabilities, believing that it would successfully assist them in their task.
    \item Users found the robot's behavior predictable and consistent, enabling them to anticipate its actions and plan their own accordingly.
    \item Users found the robot's behavior transparent and easy to understand, enabling them to anticipate and predict its actions.
\end{enumerate}

\OUT{\textcolor{red}{The average Likert of 5.89 for the first group of questions indicates that (1) users felt physically and emotionally at ease while interacting with the system, (2) users perceive the robot as a suitable and acceptable assistant for the task, and (3) users perceived the interaction with the robot as natural and seamless, resembling human-human interaction. The average Likert of 5.69 for the second group of questions indicates that (1) users perceived the robot as attentive and responsive to their needs and commands, (2) users perceived the robot as proactive and capable of autonomously driving the interaction when necessary, and (3) users perceived the robot's behavior as natural and human-like, making the interaction more intuitive. The average Likert of 6.08 for the third group of questions suggests that (1) users had high levels of trust in the robot's capabilities, believing that it would successfully assist them in their task, (2) users found the robot's behavior predictable and consistent, enabling them to anticipate its actions and plan their own accordingly, (3) users found the robot's behavior transparent and easy to understand, enabling them to anticipate and predict its actions.}}
\bigskip
The average Likert of 4.33 for the fourth group of questions suggests that users generally found it easier to interact with the robot over time and adapted their behavior during the interactions. This indicates that the system supports user learning and adaptation, making the interaction more intuitive and user-friendly with repeated use. The average Likert of 1.69 for the fifth group of questions indicates a positive outcome, as a lower score is desirable.  That means users generally did not feel tired, frustrated, or bored during the interaction trials. This indicates that the system effectively maintained user engagement and provided a positive, low-stress interaction experience.

Finally, based on the high average Likert of 4.25 for the last question, users generally rated their experience with the robot as very positive, exceeding their expectations. This suggests that the system is well-received by users and is likely to be considered effective and reliable for its intended purpose.

\section{Discussion}
\label{sec: Discussion}
As mentioned in the previous section, the performance of our RL-based system has significantly improved compared to the system in~\cite{8968505}. However, this comparison is not entirely accurate due to the evaluations being conducted on two different datasets in the two studies. In this section, we investigate this improvement in depth. Similar to Section~\ref{subsection: quant-results}, we evaluated the overall performance of the HBATNs-based system from~\cite{8968505}, as well as measuring the performance of each of its components independently using the data collected in our current human study.

Evaluating the entire HBATNs-based system, we observed 10.23\% (48 out of 469) non-eligible/non-understanding responses. We also evaluated the performance of the HBATNs-based system's STT component, finding that the percentage of SSREs in this case is 17.34\%. In contrast, the SSREs in our RL-based system's STT component are only 6.4\%, demonstrating a significant improvement. This enhancement is attributed to the inclusion of an NLTK-based dictionary, as explained in~\ref{subsection: Implementation}.

We evaluated the BERT-based DA classifier of the HBATNs-based system, finding that it classified 17.69\% of the DAs incorrectly. In comparison, the ALBERT-based model in our RL-based system classified 15.14\% of the DAs incorrectly, demonstrating that they perform almost the same in this task. However, one major difference is that the BERT-based classifier depends heavily on the context of the task and the history of the interaction. If the BERT-based DA classifier were used on a different task, it would perform poorly. In contrast, the ALBERT-based classifier developed in the current study is more generic, independent of the task, and does not rely on the history of the interaction. This robustness is due to the data augmentation and the use of single sentences as input features for training our DA classifier.

For pointing gesture recognition, since we used the exact same model as the one in the previous study, we can directly compare the percentage of incorrect pointing gestures. This percentage decreased from 28.9\% to 13.64\% solely by engineering a more accurate camera calibration.

Passing the input data through the Perception Module of the HBATNs-based system yields their corresponding Perceived Human Intent vectors. Among these, 455 were valid vectors. Subsequently, after feeding them to the Interaction Manager (HBATNs model) of the HBATNs-based system, we observed that the HBATNs model was unable to provide a proper response to 14 of them, resulting in an accuracy of 96.92\%. This deficiency stemmed from the omission of some DA-state pairs during the construction of the HBATNs model. As previously stated, there were a total of 48 non-understandings. Of these, 14 were attributed to errors within the HBATNs model, accounting for 29.17\% of the non-understandings, while the remaining 70.83\% were due to errors in other components. In contrast, only 15.22\% of non-eligible/non-understanding responses were attributable to the RL policy in our RL-based experiment.

\setlength{\tabcolsep}{1pt}
\begin{table}[t]
\caption{RL-based and HBATN-based systems comparison}\label{table: systems-comparison}
\footnotesize
\begin{tabular*}{\textwidth}{@{\extracolsep\fill}lccccccccc@{}}
\toprule
System& \cellcolor{gray!40}Avg. &\cellcolor{gray!40}Success&Non-elig.&\cellcolor{red!40}SSREs&\cellcolor{red!40}Wrong&\cellcolor{red!40}Wrong&\cellcolor{green!40}Interaction    &\cellcolor{green!40}Int. Man. Cont.&Overall\\
      &\cellcolor{gray!40}\#Moves&\cellcolor{gray!40}Rate   &Actions &\cellcolor{red!40}     &\cellcolor{red!40}DAs  &\cellcolor{red!40}Pointings&\cellcolor{green!40}Manager Acc.&\cellcolor{green!40}to Non-elig.&Likret\\
\midrule
RL-based & \cellcolor{gray!40}12.6 & \cellcolor{gray!40}96\% & 9.8\% & \cellcolor{red!40}6.4\% & \cellcolor{red!40}15.14\% & \cellcolor{red!40}13.64\% & \cellcolor{green!40}97.5\% & \cellcolor{green!40}15.22\% & 4.25 \\
HBATNs-based & \cellcolor{gray!40}15.6 & \cellcolor{gray!40}85.7\% & 10.23\% & \cellcolor{red!40}17.34\% & \cellcolor{red!40}17.69\% & \cellcolor{red!40}28.9\% & \cellcolor{green!40}96.92\% & \cellcolor{green!40}29.17\% & 4 \\
\bottomrule
\end{tabular*}
\end{table}

Finally, we improved state handling at the end of trials when the agent finds the object of interest. This ensures that the agent terminates the interaction once it locates the object. Table~\ref{table: systems-comparison} provides a summary of the comparisons between the two systems. The table also illustrates that the overall quality of the system, as rated by users on a Likert scale, has improved from 4 to 4.25 based on their survey responses. It's worth noting that the overall quality was the only aspect we could directly compare, as the HBATNs-based experiment included only that single question in their survey. However, to provide a more comprehensive analysis, we conducted a more detailed survey aiming to gather a broader range of feedback, as explained in~\ref{subsection: survey-results}.

\section{Conclusion}
\label{sec: Conclusion}
In conclusion, this paper presents a significant advancement in the field of assistive robotics, particularly in the development of interaction managers for collaborative tasks involving older adults and individuals with disabilities. By proposing a reinforcement learning approach, we address the limitations of handcrafted policies and data sparsity encountered in traditional methods. Our RL-based interaction manager, trained with a simulator using human data, demonstrates promising results in multimodal interactions, effectively handling speech, gestures, and physical actions. Through a comprehensive evaluation, we highlight the superiority of our RL-based system over the HBATNs-based approach, particularly in terms of speech recognition accuracy, DA classification performance, and overall system quality. Notably, our system achieves a high level of user satisfaction, as evidenced by positive feedback from a human study. By leveraging RL, we provide a scalable solution for building interaction managers that can adapt to various tasks and environments. Future work may focus on further refining our RL framework, exploring additional applications, and generalizing it to other tasks. Overall, this research contributes to advancing the state-of-the-art in assistive robotics and lays the foundation for more intelligent and responsive robotic assistants in the future.

\section*{Declarations}

\subsection*{Funding}
This work has been supported by the National Science Foundation grants IIS-1705058 and CMMI-1762924.

\subsection*{Conflict of interest/Competing interests}
The authors have no relevant financial or non-financial interests to disclose.

\subsection*{Ethics approval and consent to participate}
\begin{enumerate}
\item \textbf{Approval Statement:} All experimental protocols were approved by the University of Illinois Chicago Institutional Review Board (IRB).
\item \textbf{Accordance Statement:} The methods described in this study were carried out in accordance with the guidelines and regulations provided by the University of Illinois Chicago Institutional Review Board (IRB).
\item \textbf{Informed Consent Statement:} Informed consent was obtained from all participants involved in experiments using the University of Illinois Chicago Institutional Review Board (IRB) Social, Behavioral, and Educational Research Informed Consent Template.
\end{enumerate}

\subsection*{Consent for publication} All listed authors have approved the manuscript before submission, including the names and order of authors.

\subsection*{Data Availability}
The data that support the findings of this study are available from the corresponding author upon reasonable request.

\subsection*{Materials Availability}
Materials and tools used in this study are available upon request from the corresponding author.

\subsection*{Code Availability}
The custom code used in this study is available upon request from the corresponding author.

\subsection*{Author contribution:} All authors contributed to the study conception and design. Material preparation, data collection, and analysis were performed by Afagh Mehri Shervedani and Siyu Li. The first draft of the manuscript was written by Afagh Mehri Shervedani, and all authors commented on previous versions of the manuscript. All authors read and approved the final manuscript.

\begin{appendices}

\section{The survey questions}\label{sec: Appendix}
The survey questions are grouped as follows.

\begin{enumerate}
    \item \textbf{Comfort:}
    \begin{itemize}
        \item How comfortable were you during the interaction? (7-point Likert, 1:Very Uncomfortable, 2:Uncomfortable, 3:Slightly Uncomfortable, 4:Neutral, 5:Slightly Comfortable, 6:Comfortable, 7:Very Comfortable)
        \item If you were to have an assistant to help you find objects from various locations, would it be acceptable to have this robot as your assistant? (under the condition that the places are reachable for the robot) (7-point Likert, 1:Very Unacceptable, 2:Unacceptable, 3:Slightly Unacceptable, 4:Neutral, 5:Slightly Acceptable, 6:Acceptable, 7:Very Acceptable)
        \item How well was the robot able to make the interaction natural and relaxed? (7-point Likert, 1:Very Poor, 2:Poor, 3:Slightly Poor, 4:Neutral, 5:Well, 6:Slightly Well, 7:Very Well)
    \end{itemize}
    
    \item \textbf{Responsiveness:}
    \begin{itemize}
        \item How well was the robot able to respond to your actions during the interaction? (7-point Likert, 1:Very Poor, 2:Poor, 3:Slightly Poor, 4:Neutral, 5:Well, 6:Slightly Well, 7:Very Well)
        \item How well was the robot able to take the initiative and move the interaction forward when the instructions weren't clear? (7-point Likert, 1:Very Poor, 2:Poor, 3:Slightly Poor, 4:Neutral, 5:Well, 6:Slightly Well, 7:Very Well)
        \item How similar was the robot’s behavior to how a human assistant would behave? (7-point Likert, 1:Very Unsimilar, 2:Unsimilar, 3:Slightly Unsimilar, 4:Neutral, 5:Slightly Similar, 6:Similar, 7:Very Similar)
    \end{itemize}
    
    \item \textbf{Predictability:}
    \begin{itemize}
        \item How confident were you that the robot would be able to help you find the object you were looking for? (7-point Likert, 1:Very Unconfident, 2:Unconfident, 3:Slightly Unconfident, 4:Neutral, 5:Slightly Confident, 6:Confident, 7:Very Confident)
        \item How well were you able to predict how the robot would act during the interaction? (7-point Likert, 1:Very Poor, 2:Poor, 3:Slightly Poor, 4:Neutral, 5:Well, 6:Slightly Well, 7:Very Well)
        \item How well was the robot able to express what it was trying to do? (7-point Likert, 1:Very Poor, 2:Poor, 3:Slightly Poor, 4:Neutral, 5:Well, 6:Slightly Well, 7:Very Well)
    \end{itemize}

    \item \textbf{Familiarity, Learning:}
    \begin{itemize}
        \item As you progressed through the trials, did you find it easier to interact with the robot? (5-point Likert scale, 1:Strongly Disagree, 2:Disagree, 3:Neutral, 4:Agree, 5:Strongly Agree)
        \item As you progressed through the trials, did you adapt your behavior? (5-point Likert scale, 1:Strongly Disagree, 2:Disagree, 3:Neutral, 4:Agree, 5:Strongly Agree)
    \end{itemize}

    \item \textbf{Fatigue, Habituation:}
    \begin{itemize}
        \item How tired did you get as trials went on? (5-point Likert scale, 1:Not tired at all, 2:Slightly tired, 3:Moderately tired, 4:Tired, 5:Significantly tired)
        \item How frustrated did you get as trials went on? (5-point Likert scale, 1: Not frustrated at all, 2:Slightly frustrated, 3:Moderately frustrated, 4:Frustrated, 5: Significantly frustrated)
        \item How bored did you get as trials went on? (5-point Likert scale, 1: Not bored at all, 2:Slightly bored, 3:Moderately bored, 4:Bored, 5: Significantly bored)        
    \end{itemize}

    \item \textbf{Overall Quality:}
    \begin{itemize}
        \item How would you rate your experience according to your expectations? (5-point Likert scale, 1: Significantly worse than expected, 2: Worse than expected, 3: Same as expected, 4: Better than expected, 5: Significantly better than expected)
    \end{itemize}
    
\end{enumerate}

\end{appendices}

\bibliography{references}

\end{document}